\def\year{2020}\relax
\documentclass[letterpaper]{article} 
\usepackage{aaai20}  
\usepackage{times}  
\usepackage{helvet} 
\usepackage{courier}  
\usepackage[hyphens]{url}  
\usepackage{graphicx} 
\usepackage{graphicx}  
\usepackage{fancyvrb}
\usepackage{array}

\usepackage{multirow}

\usepackage{amsmath}
\usepackage{mathtools}
\usepackage{amssymb}
\usepackage[ruled, vlined,linesnumbered]{algorithm2e}
\usepackage{array, makecell}
\setcellgapes{1pt}

\usepackage{tikz}
\usetikzlibrary{fit}
\usetikzlibrary{backgrounds}
\usetikzlibrary{automata}
\usetikzlibrary{shapes}
\usetikzlibrary{matrix}
\usetikzlibrary{fit}
\usetikzlibrary{calc}
\usetikzlibrary{positioning}
\usetikzlibrary{intersections}

\tikzstyle{Player1}=[circle, thick, minimum size=0.6cm, inner 
sep=0cm,draw=black]
\tikzstyle{State}=[circle, font = \small, thick, minimum size=0.6cm, inner 
sep=0cm,draw=black,fill=white]
\tikzstyle{Final}=[circle, accepting, thick, minimum size=0.6cm, inner 
sep=0cm,draw=black]
\tikzstyle{RState}=[circle, very thick, minimum size=0.8cm, inner 
sep=0cm,draw=red]
\tikzstyle{tran}=[draw,->,font=\small]
\tikzstyle{obstyle}=[rounded corners,fill=gray!20]

\newcolumntype{H}{>{\lrbox0}c<{\endlrbox}@{}}

\newcommand{\llIf}[2]{{\let\par\relax\lIf{#1}{#2}}}
\newcommand{\llElse}[1]{{\let\par\relax\lElse{#1}}}

\newtheorem{definition}{Definition}

\newtheorem{example}{Example}

\newcommand{\tstep}{i}

\newcommand{\mdp}{\mathcal{M}}
\newcommand{\states}{\mathcal{S}}

\newcommand{\act}{\mathcal{A}}
\newcommand{\trans}{\delta}

\newcommand{\reward}{\mathit{rew}}

\newcommand{\distr}{\mathcal{D}}

\newcommand{\reals}{\mathbb{R}}
\newcommand{\last}{\mathit{last}}

\newcommand{\discount}{\gamma}

\newcommand{\hist}{h}

\newcommand{\histfunc}{H}

\newcommand{\E}{\mathbb{E}}

\newcommand{\thr}{\Delta}
\newcommand{\Rset}{\mathbb{R}}

\newcommand{\len}[1]{\mathit{len}(#1)}

\newcommand{\risk}{\mathit{Risk}}
\newcommand{\probm}{\mathbb{P}}

\newcommand{\rew}{\reward}
\newcommand{\node}{n}
\newcommand{\tree}{\mathcal{T}}

\newcommand{\payoff}{\mathit{Payoff}}
\newcommand{\fail}{F}
\newcommand{\pred}{f}
\newcommand{\param}{\theta}
\newcommand{\valpred}{v}
\newcommand{\riskpred}{r}
\newcommand{\probpred}{\mathbf{p}}
\newcommand{\episodes}{\mathit{episodes}}
\newcommand{\batch}{\mathit{batch}}
\newcommand{\trset}{\mathit{Data}}
\newcommand{\algomode}{\mathit{mod}}
\newcommand{\traindata}{E}
\newcommand{\stepdist}{\xi}
\newcommand{\trroot}{\mathit{root}}
\newcommand{\riskdist}{\mathit{\tau}}
\newcommand{\explconst}{C}
\newcommand{\propval}{\mathit{val}}

\newcommand{\linprog}{\mathcal{L}}
\newcommand{\leaf}{\mathit{leaf}}
\newcommand{\flow}{\mathit{Flow}}
\newcommand{\mypar}[1]{\smallskip\noindent\textbf{#1}.}

\newcommand{\lrate}{\alpha}
\newcommand{\hor}{H}

\newcommand{\ralph}{RAlph}
\newcommand{\epi}{\eta}
\newcommand{\steprew}{\rho}
\newcommand{\simnum}{\mathit{sim}}

\newcommand{\expnodes}{\simnum}

\title{Reinforcement Learning  of Risk-Constrained Policies\\ in Markov Decision Processes}
\author{\Large \textbf{Tom\'a\v{s} Br\'{a}zdil\textsuperscript{\rm 1}, Krishnendu Chatterjee\textsuperscript{\rm 2}, Petr Novotn\'y\textsuperscript{\rm 1}, Ji\v{r}\'i Vahala\textsuperscript{\rm 1}}\\ 
\textsuperscript{\rm 1}Faculty of Informatics, Masaryk University, Brno, Czech Republic\\ 
\{xbrazdil, petr.novotny, xvahala1\}@fi.muni.cz \\ 
\textsuperscript{\rm 2}Institute of Science and Technology Austria, Klosterneuburg, Austria\\ 
Krishnendu.Chatterjee@ist.ac.at 
}

\begin{document}

\maketitle

\begin{abstract}
Markov decision processes (MDPs) are the defacto framework for sequential 
decision making in the presence of stochastic uncertainty. 
A classical optimization criterion for MDPs is to maximize the expected 
discounted-sum payoff, which ignores low probability catastrophic events with 
highly negative impact on the system.
On the other hand, risk-averse policies require the probability of 
undesirable events to be below a given threshold, but they do not 
account for optimization of the expected payoff.
We consider MDPs with discounted-sum payoff with failure states which 
represent catastrophic outcomes.
The objective of \emph{risk-constrained} planning is to maximize the expected discounted-sum payoff among 
risk-averse policies that ensure the probability to encounter a failure 
state is below a desired threshold.
Our main contribution is an efficient risk-constrained planning algorithm that 
combines UCT-like search with a predictor learned through 
interaction with the MDP (in the style of AlphaZero) and with a risk-constrained action selection via linear programming.
We demonstrate the effectiveness of our approach with experiments 
on classical MDPs from the literature, including benchmarks with an order of $ 10^6 $ states.
\end{abstract}

\section{Introduction}

\noindent{\em MDPs with discounted-sum objectives.}
A classical problem in artificial intelligence is sequential decision 
making under uncertainty.
The standard model incorporating both decision-making choices and stochastic 
uncertainty are Markov decision processes (MDPs)~\cite{Howard,Puterman:book}.
MDPs have a wide range of applications, from 
planning~\cite{RN10}, to reinforcement learning~\cite{LearningSurvey}, 
robotics~\cite{KGFP09}, and verification of probabilistic systems~\cite{BK:book}, 
to name a few.
The objective in decision making under uncertainty is to optimize a payoff 
function.
A fundamental payoff function is the {\em discounted-sum payoff}, 
where every transition of the MDP is assigned a reward, and for an infinite path 
(that consists of an infinite sequence of transitions) the payoff is the 
discounted-sum of the rewards of the transitions.

\smallskip\noindent{\em Expectation optimization and risk.}
In the classical studies of MDPs with discounted-sum payoff the objective 
is to obtain policies that maximize the expected payoff. 
However, this ignores that low probability failure events can have 
highly negative impact on the system.
In particular, in safety critical systems, or systems with high cost for failures,
policies with high expected reward can be associated with risky actions 
with undesirable chances of failure.

\smallskip\noindent{\em CCMDPs and risk-reward tradeoff.} 
Chance- (or risk-) constrained MDPs (CCMDPs) introduce {\em chance constraint} 
or {\em risk bound} which provides a bound on the allowed probability of 
failure of a policy~\cite{Rossman77,STW16:BWC-POMDP-state-safety,Vulcan}. 
In particular, we consider MDPs equipped with a set of failure states which represent 
catastrophic outcomes. 
The probability to encounter any failure state represents the risk. 
Given a desired probability threshold for the risk bound, a risk-averse policy ensures 
that the probability of failure does not exceed the given bound.
On one hand, policies with low-risk may ensure little expected payoff; 
on the other hand, policies with high expected payoff can be associated with high risk.
Thus the relevant question to study is the interplay or the tradeoff of risk and expected payoff.
In this work we study the following {\em risk-constrained} planning problem: 
given a risk bound, the objective is to maximize the expected payoff among all risk-averse 
policies that ensure the failure probability is at most the risk bound.

\smallskip\noindent{\em Motivating scenarios.} 
Risk-constrained planning is natural in several scenarios.
For example, in planning under uncertainty (e.g., autonomous driving)
certain events (e.g., the distance between two cars, or the 
distance between a car and an obstacle, being less than a specified safe distance)
must be ensured with low probability.
Similarly, in scenarios such as a robot exploring an unknown environment 
for natural resources a significant damage of the robot ends the mission,
and must be ensured with low probability.
However, the goal is to ensure effective exploration within the specified risk
bounds, which naturally gives rise to the risk-constrained planning problem we
consider.

\smallskip\noindent{\em Our contributions.} 
The risk-constrained planning problem (or CCMDPs) have been considered in 
previous works such as~\cite{STW16:BWC-POMDP-state-safety,Vulcan}. 
However these works consider only deterministic policies, and 
randomized (or mixed) policies are strictly more powerful for the 
risk-constrained planning problem~\cite{Altman:book}. 
A possible approach for the risk-constrained planning problem is via 
linear programming or dynamic programming methods, however, they scale poorly and
are unsuitable for large state spaces~\cite{Vulcan}. 
Our main contribution is an efficient risk-constrained planning algorithm that 
combines UCT-like search with a predictor learned through 
interaction with the MDP and with a risk-constrained 
action selection via linear programming over a {\em small sampled} tree-shaped MDP.
Since the linear programming is over a sampled sub-MDP, 
our algorithm is scalable as compared to linear programming over the entire MDP, while the use of predictor significantly enhances the search. By using the predictor we lose formal guarantees on the solution, but gain in performance. We also show that despite the lack of guarantees, our method converges to well-behaved policies in practice.
We demonstrate this with experiments 
on classical MDPs from the literature, including benchmarks with an order of $ 10^6 $ 
states.

\smallskip\noindent \textbf{Related Work.}
Discounted-payoff MDPs are a well-established model~\cite{Puterman:book,FV97}.
The notion of ensuring risk constraints is also well-studied~\cite{Rossman77,HYV16:risk-pomdps}.
Moreover, CCMDPs can be considered as a special case of constrained MDPs (CMDPs)~\cite{Altman:book}.
CMDPs are often solved using linear programming approaches which do not scale to 
large MDPs~\cite{Vulcan}.
The works most closely related to the problem we 
consider are as follows:
First, the risk-constrained planning for partially-observable MDPs (POMDPs) 
with deterministic policies has been considered in~\cite{STW16:BWC-POMDP-state-safety},
and risk-constrained MDPs with deterministic policies have been considered in~\cite{Vulcan}.
In contrast, we consider randomized policies, which are 
 more powerful for risk-constrained planning.
Another related approach for POMDPs are \emph{constrained POMDPs}~\cite{UH10:constrained-pomdp-online,PMPKGB15:constrained-POMDP}, 
where the objective is to maximize the expected payoff ensuring that the expected payoff of 
another quantity is bounded.
Risk-constrained MDP optimization with randomized policies was considered in \cite{TK12}. There they consider optimization under formally guaranteed PCTL constraints via an iterative linear programming (LP) over the whole state space. The largest benchmark reported in the referenced paper has $ 75^2 $ states, while we report MDPs with up to ca. $ 6.5\cdot 10^6 $ states. Hence, the method of \cite{TK12} is preferable where guarantees are a priority while RAlph is preferable where scalability is a priority. The paper~\cite{BTT18} considers \emph{stochastic shortest path} under PLTL constraints, i.e. the  rewards are positive costs and we minimize the expected cost of reaching a target. In contrast, we consider arbitrary rewards under safety constraints.

Several problems related to risk-constrained planning with other objectives have been 
considered, such as:
(a)~risk threshold~0 for long-run average and stochastic shortest path problems MDPs~\cite{bfrr14,rrs15}; 
(b)~general risk threshold for long-run average payoff in MDPs~\cite{ckk15}. 
(c)~risk bound~0 for discounted-sum POMDPs~\cite{CNPRZ17:BWC-POMDP}; and
(d)~general risk bound for discounted-sum  POMDPs~\cite{CENR:ijcai18}.
In all these works the risk is formulated as risk of the payoff being below a given value rather than
of reaching failure states.
Moreover, these works (apart from d)) focus on dynamic programming methods, rather than scalable algorithms
for large MDPs. Although d) also uses linear programming over a sampled sub-MDP, it does not use predictors and its tree-search procedure is closer to the original UCT~\cite{KS06} than to its more sophisticated version used by AlphaZero~\cite{alphagozero,alphazero-science}. While the algorithm of d) can be adapted to risk-constrained MDPs with reachability risk, our experiments show that our new algorithm scales much better.

\section{Preliminaries}


\begin{definition}
A \emph{Markov decision process (MDP)} is a
	tuple $\mdp=(\states,\act,\trans,\reward,s_0,\discount)$ 
	where
	$\states$ is a  set of \emph{states},
	$\act$ is a set of \emph{actions},
	$\trans:\states\times\act \rightarrow \distr(\states)$ is a 
	\emph{probabilistic transition function} that given a state $s\in \states$ and an
	action $a\in \act$ gives the probability distribution over the successor 
	states,
	$\reward: \states \times \act \rightarrow \reals$  is a \emph{reward 
	function}, $s_0$ is the \emph{initial state}, and $\discount\in(0,1]$ is the \emph{discount factor}.
	We abbreviate $\trans(s,a)(s')$ by 
	$\trans(s'|s,a)$. 
\end{definition}

\mypar{Policies}
The interaction with an MDP starts in the initial state $s_0$ and proceeds sequentially through a \emph{policy} $\pi$, a computable function which acts as a blueprint for selecting actions, producing longer and longer \emph{history} of actions and observations. Formally, a history is an alternating sequence of states and actions starting and ending with a state. The initial history is $\histfunc_0 = s_0$. In every time step $\tstep\in \{0,1,2,\ldots\}$ the interaction already produced some history $\histfunc_{\tstep}$ whose last state $\last(\histfunc_\tstep)$ is the current state $S_\tstep$ of the system. In such a situation, $\pi$ selects an action $A_\tstep \in \act$ to play in step $\tstep$. The choice may depend on the whole past history, and it might also be  randomized, i.e. $A_\tstep \sim \pi(\histfunc_\tstep)$. The agent then gets an immediate reward $ \mathit{Rew}_\tstep = \reward(S_\tstep,A_\tstep) $ and proceeds to the next state $S_{\tstep + 1}$, which is sampled according to the transition function, i.e. $S_{\tstep +1} \sim \trans(S_{\tstep},A_{\tstep})$. Thus, the current history is now $ \histfunc_{\tstep+1}= \histfunc_\tstep A_\tstep S_{\tstep+1}$, obtained from the previous history by appending the last selected action and the resulting state. Throughout the text we denote by $ S_\tstep,A_\tstep,\histfunc_\tstep $ the random variables returning the state, action, and current history in step $ \tstep $, while the notation $ s,a,\hist $, etc. is reserved for concrete states/actions/histories (i.e. elements of the co-domains of $ S_\tstep/A_\tstep/\histfunc_\tstep $). 

We denote by $\probm^\pi_{}(E)$ the probability of an event $ E $ under policy $\pi$, and by $ \E^\pi_{}[X] $ the expected value of a random variable $ X $ under $\pi$. 

\mypar{Payoffs}
The expected payoff of a policy $\pi$ from state $s$ is the value $\payoff(\pi,s) = \E^\pi_s[\sum_{\tstep=0}^{\infty} \discount^\tstep \cdot \mathit{Rew}_{\tstep}]$. 

\mypar{Risk-Constrained Optimization}
To encompass the notion of an undesirable event, we equip each MDP $\mdp=(\states,\act,\trans,\reward,s_0,\discount)$ with a set $\fail_\mdp \subseteq \states$ of \emph{failure states}. A \emph{risk} of a policy $\pi$ is then the probability that a failure state is encountered: 
$ \risk(\pi) = \probm^\pi_s\Big( \bigcup_{\tstep=0}^{\infty}\{S_\tstep \in \fail_\mdp \} \Big).
 $
We assume that each $s \in \fail_\mdp$ is a \emph{sink,} i.e. $ \trans(s|s,a) = 1 $ and $ \reward(s,a)=0 $ for all $ a\in\act $. Hence, $ \fail_\mdp $ models failures after which the agent has to cease interacting with the environment (e.g. due to being destroyed).

The risk-constrained planning problem is defined as follows: given an MDP $ \mdp $ and a \emph{risk threshold} $ \thr \in [0,1]$, find a policy $ \pi $ which maximizes $ \payoff(\pi) $ subject to the constraint that $ \risk(\pi) \leq \thr $. If there is no \emph{feasible} policy, i.e. a policy s.t.  $ \risk(\pi) \leq \thr $, then we want to find a policy that minimizes the risk and among all such policies optimizes the expected payoff.

In this paper, we present \ralph{} (a portmanteau of ``Risk'' and ``Alpha''), an online algorithm for risk-constrained planning. Inspired by the successful approach of AlphaZero, \ralph{} combines a UCT-like tree search with evaluation of the leaf nodes via a suitable \emph{predictor} learned through a repeated interaction with the system. On top of this, we augment the algorithm's action-selection phase with a risk-constrained mechanism based on evaluation of a linear program over the constructed search tree. 


\section{The Algorithm}

\mypar{Predictor} First we formally define the notion of a predictor. A predictor is a $\param$-parameterized function $ \pred_{\param} \colon \states \rightarrow \Rset\times[0,1]\times [0,1]^{|\act|}$ assigning to each state $s$ the tuple 
$\pred_{\param}(s)=(\valpred,\riskpred,\probpred)$ 
 which predicts the parameters of some policy $\pi$:
 $\valpred$ is the predicted expected payoff of $\pi$ from $s$,
 $\riskpred$ is the predicted risk of $\pi$ from $s$, and 
 $\probpred$ is the vector of \emph{prior probabilities} over the set $\act$ in $s$.
%
We defer the details of the predictor implementation, its parameters, and the learning technique used to update them, to Subsection~\ref{subsec:training}. 


\mypar{\ralph{}: Overall Structure} The main training and evaluation loops of \ralph{} are given in Algorithm~\ref{algo:ralphmain}. As in other algorithms based on search through the search tree, termination is ensured by searching only up to a given finite horizon $ \hor $. In the training phase, \ralph{} repeatedly samples episodes of the agent-environment interaction, using the \texttt{\ralph-episode} procedure described in Subsection~\ref{subses:rcts}. After each batch of episodes is sampled, the gathered data are used to retrain the predictor via the procedure \texttt{Train}, described in Subsection~\ref{subsec:training}. Once the training is finished, we fix the predictor and continue to the evaluation phase.

\begin{algorithm}[t]
\SetInd{0.7em}{0.3em}
	\SetKwProg{proc}{procedure}{}{}
	\SetKwFunction{ralphtrain}{\ralph-train}
	\SetKwFunction{ralpheval}{\ralph-evaluate}
	\SetKwFunction{ralphepi}{\ralph-episode}
	\SetKwFunction{retrain}{Train}
	\SetKwFunction{update}{UpdateTrees}
	\SetKwFunction{rollout}{Rollout}
	\SetKwFunction{explore}{Explore}
	\SetKwFunction{select}{SelectAction}
	\SetKwFunction{play}{PlayAction}

\proc{\ralphtrain}{
\KwIn{MDP $\mdp$ (with a horizon $\hor$), risk bound $ \thr $, no. of training episodes $ m $, batch size $ n $}

$\episodes \leftarrow 0$; $\algomode \leftarrow \text{``train''}  $; initialize $ \pred_{\param} $\;
\While{$ \episodes < m $}{
	$\batch \leftarrow 0$; $ \trset \leftarrow \emptyset $\;
	\While{$ \batch < n $}{
		$E \leftarrow \;$\ralphepi($\mdp$,$\hor$,$\pred_{\param}$,$\thr$,$\algomode$)\; 
		$\batch \leftarrow \batch+1$\; $\episodes \leftarrow \episodes + 1  $\;
		$ \trset \leftarrow \trset \cup \{E\} $\;
	}
	$\param \leftarrow \;$ \retrain($ \param $, $\trset$)
}
}

\proc{\ralpheval}{
\KwIn{MDP $\mdp$ (with a horizon $\hor$), risk bound $ \thr $, pre-trained predictor $ \pred_{\param} $}
$\algomode \leftarrow \text{``eval''}$\;
\lWhile{true}{
	\ralphepi($\mdp$,$\hor$,$\pred_{\param}$,$\algomode$)\DontPrintSemicolon 
}
}
\caption{Training and evaluation of \ralph.}
\label{algo:ralphmain}
\end{algorithm}


\subsection{Risk-Constrained Tree Search}
\label{subses:rcts}

In this subsection we describe the procedure \texttt{\ralph-episode} (Algorithm~\ref{algo:ralphepi}). We first describe the conceptual elements of the algorithm, then the data structures it operates on, and finally the algorithm itself.

\mypar{Overview} The algorithm interacts with the MDP for $\hor$ steps, each step $\tstep$ resulting in the (randomized) choice of some action $a_\tstep$ to be played. 
In every step, \ralph{} first expands the search tree $\tree$ by iterating the usual 4-phase MCTS simulations (node selection, tree expansion, leaf evaluation, backpropagation, see procedure \texttt{Simulate}). 
We follow the spirit of \cite{alphazero-science} and use the predictor $\pred_{\param}$ to evaluate the leaf nodes. Using the data stored within the tree, we then compute the distribution from which $ a_\tstep $ is sampled.

To accommodate the risk, we extend the AlphaZero-like MCTS with several conceptual changes, outlined below.


\textit{Risk-constrained sampling of $ a_\tstep $.} 
In the action selection phase, we solve a linear program (LP) over $ \tree $, which yields a local policy that maximizes the estimated payoff while keeping the estimated risk below the threshold $\thr$ (line~\ref{aline:flow}; described below). The distribution $ \stepdist_\tstep $ used by the local policy in the first step is then used to sample $a_\tstep$. 

\textit{Risk-constrained exploration.} Some variants of AlphaZero enable additional exploration by  selecting each action with a probability proportional to its exponentiated visit count~\cite{alphagozero}. Our algorithm use a technique which perturbs the distribution computed by the LP while keeping the risk estimate of the perturbed distribution below the required threshold (line~\ref{aline:explore}; described below).

\textit{Risk predictor.} In our algorithm, the predictor is extended with risk prediction.
 
\textit{Estimation of alternative risk.} The risk threshold must be updated after playing an action, see Example~\ref{ex:riskup}, since each possible outcome of the action has a potential contribution towards the global risk. We use linear programming and the risk predictor to obtain an estimate of these contributions.



\mypar{Data Structures} 
The search tree~\cite{SV:POMCP}, denoted by $\tree$, is a dynamic tree-like data structure whose nodes correspond to histories of $\mdp$. We name the nodes directly by the corresponding histories. Each child of a node $\hist$ is of the form $\hist a t$, where $a\in \act$ and $t \in \states$ are s.t. $\trans(t|\last(\hist),a)>0$. Each node $\hist$ has these attributes: 
\begin{itemize}
\item
$\hist.N$, the visit count of $\hist$;
\item $\hist.\valpred$ and $\hist.\riskpred$; the last predictions of payoff and risk obtained by $\pred_{\param}$ for $\last(\hist)$. 
\end{itemize}
Moreover, for each action $a\in \act$ we have the attributes:
\begin{itemize}
\item 
 $\hist. N_a$, counting the number of uses of $a$ during visits of $\hist$;
 \item
 $ \hist. V_a $, the average payoff accumulated by past simulations after using $a$ in $\hist$; 
 \item 
 $ \hist.p_a $, the last prediction of a prior probability of $a$ obtained by $\pred_{\param}$ in $\last(\hist)$.
\end{itemize}
We also have the following derived attributes:
 $\hist.V_{\min} = \min_{ a\in \act} \hist.V_a$; and 
 $\hist.V_{\max} = \max_{a\in\act} \hist.V_a$.
These are re-computed to match the defining formulae whenever some $\hist.V_a$  is changed. Every newly created node is initialized with zero attributes.

We denote by $\trroot(\tree)$ the root of $\tree$ and by $\leaf(\tree)$ the set of leafs of $ \tree $. Also, for a node $ \hist $ we denote by $ \tree(\hist) $ the sub-tree rooted in $ \hist $.

\mypar{Episode Sampling: Overall Structure}  In Algorithm~\ref{algo:ralphepi}, a single search tree $ \tree $ is used as a global dynamic structure. In this paragraph, we provide a high-level description; the following paragraphs contain details of individual components of the algorithm. The main loop (lines \ref{aline:mloopbeg}--\ref{aline:mloopend}) has a UCT-like structure. In every decision step $ \tstep $, the tree is extended via a sequence of simulations (described below); the number of simulations being either fixed in advance or controlled by setting a timeout. After that, we solve a linear program over $ \tree $ defined below (line \ref{aline:flow}). This gives us a distribution $ \stepdist_\tstep $ over actions as well as a risk distribution $\riskdist_i$ over the child nodes of the root node $\nu$. Informally, $\riskdist(\nu b t)$ is the estimated future risk of hitting a failure state after playing $ b $ and transitioning into $ t $. After solving the program, we sample an action $a_\tstep$ to play and then the corresponding successor state $s_{\tstep+1}$, obtaining an immediate reward $\steprew_i$. The risk threshold is then updated to by the formula on lines~\ref{aline:thrupbeg}--\ref{aline:thrupend}, where $ \mathit{altrisk} $ is the probability placed by the risk distribution on all the histories not consistent with the current history $\nu a_i s_{i+1}$. Finally, we prune away all parts of the tree not consistent with the current history and continue into a next iteration.


\mypar{Simulations \& UCT Selection} Simulations also follows the standard UCT scheme. In every simulation, we traverse the tree from top to bottom by selecting, in every node $ \hist $, an action  $a = \arg\max_{a\in \act} \texttt{UCT}(\hist,a) $, where
\[
\texttt{UCT}(\hist,a) = \frac{\hist.V_a - \hist.V_{\min}}{\hist.V_{\max} - \hist.V_{\min}} + \explconst \cdot \hist.p_a \cdot \sqrt{\frac{\ln (\hist.N)}{\hist.N_a + 1}}.
\]
Here $ \explconst $ is a suitable \emph{exploration constant,} a parameter fixed in advance of the computation. 

Upon reaching a leaf node $ \hist $, we expand $ \tree $ by adding \emph{all} possible child nodes of $ \hist $ (lines \ref{aline:expandbeg}--\ref{aline:expandend}). 
Finally, we perform a bottom-up traversal from $ \hist $ to the root, updating the node and action statistics with the data from the current simulation (lines~\ref{aline:dataupbeg}--\ref{aline:dataupend}). Note that the payoff and risk from $ \hist $ (which was not visited by a simulation before) is estimated via the predictor, unless $ \hist $ corresponds either to being trapped in a failure state (in which case its risk is clearly one and future payoff $ 0 $) or to running out of the horizon without hitting $ \fail_\mdp $, in which case the risk and future payoff are both $ 0 $.

\mypar{Linear Program} We first fix some notation. For a history $h = s_0 a_0 s_1 a_1 \dots a_{n-1} s_n$ we define its length $\len{h}$ to be $n$ and its payoff to be$\payoff(h) = \sum_{i=0}^{\len{h}-1}\gamma^i\cdot\rew(s_i,a_i)$. 

The procedure \texttt{Solve-LP}$(\tree,\thr)$ constructs a linear program $ \linprog$, which has variables $ x_{\hist}, x_{\hist,a} $ for every node $\hist\in\tree$ and every $a\in \act$, and is pictured in Figure~\ref{fig:lp}.
\begin{figure}[t] 
\begin{align}
&\textit{max} \sum_{\hist\in\leaf(\tree)} x_{\hist}\cdot( \payoff(\hist) + \gamma^{\len{\hist}}\cdot \hist.v  ) 
\textit{ subject to}\nonumber\\
&x_{\trroot(\tree)} = 1 \label{eq:flow-start}\\
&x_{\hist} = \sum_{a\in \act} x_{\hist,a}  \quad\quad\quad\quad\quad\quad\quad\text{for } \hist \in \tree\setminus \leaf(\tree)\label{eq:flow-start-alt}\\
&x_{\hist b t} = x_{\hist,b} \cdot \trans(t\mid \last(\hist),b)  \quad\text{for } \hist, \hist b t \in \tree\\[2mm]
& 0 \leq x_{\hist} \leq 1  ,\quad  0 \leq x_{\hist,a} \leq 1  \quad\quad \text{for } \hist \in \tree, a \in \act\label{eq:flow-end}\\
&  \sum_{\hist \in \leaf(\tree)} x_{\hist}\cdot \hist.r \leq \thr \label{eq:risk}
\end{align}
\caption{The Linear program $\linprog$.}
\label{fig:lp}
\end{figure}

The LP $ \linprog $  encodes a probabilistic flow induced by some policy (constraints \eqref{eq:flow-start}--\eqref{eq:flow-end}, which we together denote by $ \flow(\tree) $), $x_{\hist}$ being the probability that the policy produces a history $h$ and $x_{\hist,a}$ the probability that $h$ is produced and afterwards $a$ is selected. We aim to maximize the expected payoff of such a policy (with payoffs outside the tree estimated by predictions stored in $\hist.v$) while keeping the (estimated) risk below $\thr$ (constraint \eqref{eq:risk}). Hence, the procedure \texttt{Solve-LP} returns an action distribution $\stepdist_\tstep$ s.t. $ \stepdist_\tstep(a) = x_{\trroot(\tree),a} $ for each $a\in\act$. 

If $\thr=1$, there is no need for constrained sampling. Hence, in such a case we omit the LP step altogether and make the selection based on the action visit count.

\begin{example}
\label{ex:lp} 
Consider an MDP $\mdp=(\states,\act,\trans,\reward,s,\discount)$ with $\states=\{s,t,u\}$ and $ \act = \{a,b\} $ s.t. $\trans(s|s,a) = \trans(t|s,a)=\frac{1}{2}$, $ \trans(u|s,b)=1 $. The states $ t,u $ are sinks, $ \fail_\mdp=\{t\} $, and $ \rew(s,a)=1 $ (all other rewards are $ 0 $). We put $\discount=0.95$, and $\thr= 0.6$. Assume, for the sake of simplicity, that we have just one simulation per step, which, in the initial step, yields the following tree:
\begin{center}
\begin{tikzpicture}
\node[State] (root) at (0,0) {$s$};
\node[State] (lchild) at (-1,-1) {$sas$};
\node[State] (cchild) at (0,-1) {$sat$};
\node[State] (rchild) at (1,-1) {$sbu$};
\draw[-] (root) -- (lchild);
\draw[-] (root) -- (cchild);
\draw[-] (root) -- (rchild);
\end{tikzpicture}
\end{center}
Next, assume that the current predictor predicts risk $0.4$ for $s$, and $0.1$ for $u$, while the predicted payoffs are $0$ for $t,u$ and $1$ for $s$. Then $\linprog$ asks to maximize $x_{sas}\cdot 1.95 + x_{sat}$ under the following constraints: $x_s = 1$, $x_s = x_{s,a} + x_{s,b}$, $x_{sas} = 0.5\cdot x_{s,a} $, $x_{sat} = 0.5\cdot x_{s,a} $, $x_{sbu} =  x_{s,b} $, $0.4\cdot x_{sas} + x_{sat} + 0.1\cdot x_{sbu} \leq 0.6$ (and all variables in $[0,1]$).
\end{example}

\begin{algorithm}[!t]
\SetInd{0.7em}{0.3em}
	\SetKwProg{proc}{procedure}{}{}
	\SetKwFunction{ralphtrain}{\ralph-train}
	\SetKwFunction{ralpheval}{\ralph-evaluate}
	\SetKwFunction{ralphepi}{\ralph-episode}
	\SetKwFunction{optflow}{Solve-LP}
	\SetKwFunction{riskexplore}{RiskAwareExplore}
	\SetKwFunction{simulate}{Simulate}
	\SetKwFunction{retrain}{Train}
	\SetKwFunction{update}{UpdateTrees}
	\SetKwFunction{rollout}{Rollout}
	\SetKwFunction{explore}{Explore}
	\SetKwFunction{select}{SelectAction}
	\SetKwFunction{play}{PlayAction}
	\SetKwFunction{uprisk}{UpdateRisk}
	\SetKwFunction{uctselect}{UCT-Select}
	\SetKwFunction{predict}{Predict}
	\SetKw{kwglobal}{global}

\proc{\ralphepi($\mdp$,$\hor$,$\pred_{\param}$,$\thr$,$\algomode$)}{
\kwglobal $\tree$\;
initialize $\tree$ to one node $s_0$; $ \traindata \leftarrow $ empty sequence\;
\For{$\tstep \leftarrow 0 $ to $ \hor-1 $}{ \label{aline:mloopbeg}
$ \nu \leftarrow \trroot(\tree)  $; $s_\tstep \leftarrow \last(\nu)$\;
\Repeat{\textit{timeout}}{
	\simulate($\mdp$, $\hor-i$,$ \tree $) \tcp*{build $\tree$}
}
$\stepdist_\tstep, \riskdist_\tstep \leftarrow \optflow(\tree, \thr)$\;  \label{aline:flow}
\If{$mode =$ {\upshape ``train''}}{
	${\stepdist_\tstep} \leftarrow \riskexplore(\tree,\stepdist_\tstep) $ \label{aline:explore}
	}
$a_{\tstep} \leftarrow $ sample from $\stepdist_\tstep$\;
$\steprew_{\tstep} \leftarrow  \reward(s_{\tstep},a_{\tstep})$\; $s_{\tstep + 1} \leftarrow $ sample from $\trans(s_\tstep,a_\tstep)$\;
append $ (s_\tstep,\stepdist_\tstep,\steprew_\tstep) $ to $ \traindata $\;
$ \mathit{alt} \leftarrow \{\nu' \in \tree | \nu' \text{ child of } \nu \text{ s.t. } \nu' \neq \nu a_\tstep s_{\tstep+1} \} $\label{aline:thrupbeg}\;
$ \mathit{altrisk} \leftarrow \sum_{\nu ' \in \mathit{alt}} \riskdist_\tstep(\nu') $\;
$ \thr \leftarrow {(\thr - \mathit{altrisk})}/{\riskdist_\tstep(\nu a_\tstep s_{\tstep+1})} $ \label{aline:thrupend}\;
$ \tree \leftarrow $ sub-tree of $ \tree $ rooted in $ \nu a_is_{\tstep+1} $
}\label{aline:mloopend}
\Return{$ \traindata $} \label{aline:return}} 

%

\proc{$ \simulate(\mdp, \mathit{steps},\tree )$}{
$\hist \leftarrow \trroot(\tree)$; $ \mathit{depth} \leftarrow 0 $\;
\While{$ \hist $ is not a leaf of $ \tree $}{
$ a \leftarrow \arg\max_{a\in\act} \texttt{UCT}(\hist,a) $\;
$ s \leftarrow $ sample from $\trans(\last(\hist),a)$\;
$ \hist \leftarrow \hist a s $; $ \mathit{depth} \leftarrow \mathit{depth} +1 $\;
}
\If{$ \last(\hist) \not\in \fail_\mdp \wedge \mathit{depth} < \mathit{steps}$}{
\ForEach{$ b \in \act $}{\label{aline:expandbeg}
	\ForEach{$t\in \states$ s.t. $\trans(t|\last(\hist),b)>0$}{ 
		initialize a new leaf $\hist b t$, add it to $\tree$ as a child of $\hist$\label{aline:nodeexp} \;
		\predict($ \hist b t $)\label{aline:expandend}
	}
}}
\lElseIf{$ \last(\hist) \in \fail_\mdp $}{$ \hist.r \leftarrow 1$}\lElse{$ \hist.r\leftarrow 0 $}
$ \propval \leftarrow \hist.v$;
$ \hist.N \leftarrow \hist.N+1 $\;
\While{$ \hist \neq \trroot{(\tree)}$}{\label{aline:dataupbeg}
let $ \hist = \hist' b t $ where $b\in \act$, $ t \in \states $\;
$\hist'.N \leftarrow \hist'.N +1  $; $ \hist'.N_b \leftarrow \hist'.N_b +1 $\;
$ \propval \leftarrow \rew(\last(\hist'),b) + \gamma\cdot\propval  $\;
$ \hist'.V_b \leftarrow \hist'.V_b + (\propval - \hist'.V_b)/\hist'.N_b  $\;
$ \hist \leftarrow \hist' $\label{aline:dataupend}
}

}

\proc{\predict($ \hist, \pred_{\param} $)}{
$ (\hist.\valpred,\hist.\riskpred,(\hist.p_a)_{a\in\act}) \leftarrow \pred_{\param}(\last(\hist))$
}

\caption{The episode sampling of \ralph{}.}
\label{algo:ralphepi}
\end{algorithm}

\mypar{Risk Distribution} The choice of actions according to $ \stepdist_\tstep $ is randomized, as is the subsequent sample of the successor state. Each outcome of these random choices contributes some risk to the overall risk of our randomized policy.

\begin{example}
\label{ex:riskup}
 Consider $\mdp$ as in Example $\ref{ex:lp}$, with $\thr = 0.6$. If the agent selects action $a$ and the system transitions into the non-failure state $s$, the agent made the risk in the root node $s$ equal to $r_0 = \frac{1}{2}\cdot 1 + \frac{1}{2}\cdot r_1$, where $r_1$ is the probability of hitting a failure state after continuing the play from $s$. To ensure that $r_0 \leq \thr$, we must ensure $r_1 \leq 0.2$. Hence, in the next step, $\thr$ must be updated to $0.2 $.
\end{example}

 Hence, when making a step, we need to compute a risk distribution $ \riskdist_\tstep $ which assigns to each possible outcome (i.e. each child of $ \trroot(\tree) $) an estimate of its risk contribution. This distribution used to update the risk threshold $ \thr $ after a concrete outcome of the choices is observed (lines \ref{aline:thrupbeg} -- \ref{aline:thrupend}). 
 In our experiments, we use the \emph{optimistic} risk estimate, which assigns to each child $ \hist $ of the root the minimal risk achievable in the sub-tree rooted in $ \hist $ (the risk of leafs being estimated by $ \pred_\param $). Formally, we set $\riskdist_\tstep(\hist)$ to be the optimal value of a linear program $ \linprog_{\mathit{risk}}(\hist) $ with constraints $ \flow(\tree(\hist)) $ and with the objective to minimize $ \sum_{\hist' \in \leaf(\tree(\hist))} x_{\hist'}\cdot \hist'.r $.

\mypar{Infeasible LP} The linear program $ \linprog $ might be infeasible, either because there is no policy satisfying the risk threshold or because the risk estimates are too imprecise (and pessimistic). In such a case, we relax the overall risk constraint while trying to stay as risk-averse as possible. Formally, we reset $ \thr $ to be the minimal risk achievable in the current tree, i.e. the optimal value of $ \linprog_{\mathit{risk}}(\trroot(\tree)) $. (Note that $ \linprog_{\mathit{risk}}(\hist) $ is feasible for each node $ \hist $.) We then again solve $ \linprog$, which is guaranteed to be feasible under the new $ \thr $.

\mypar{Exploration} The exploration-enhancing procedure \texttt{RiskAwareExplore} (line \ref{aline:explore}) uses a pre-set function $\mathit{expl}$ which, given an integer $j$, returns a value from $[0,1]$. When called, the procedure   performs a Bernoulli trial with parameter $\mathit{expl}(j)$, where $j$ is the number of past calls of the procedure. Depending on outcome, it either decides to not explore (entailing no change of $\stepdist_\tstep$); or to explore, in which case we modify $ \stepdist_\tstep $ in a way depending on whether the computation of $ \stepdist_\tstep $ required just one call of the linear solver (i.e. if $ \linprog $ was feasible without relaxing $ \thr $) or not. 

If $ \linprog $ was feasible on the first try, we perturb $ \stepdist_\tstep $ using the standard Boltzmann (softmax) formula~\cite{LearningSurvey}, i.e. the perturbed probabilities are proportional to an exponential function of the original probabilities. The perturbed distribution $\tilde{\stepdist_\tstep}$ might be too risky, which is indicated by violating the risk constraint $ \sum_{b\in \act, t \in \states} \riskdist_\tstep(\trroot(\tree)bt) \cdot \tilde{\stepdist_\tstep}(\trroot(\tree)bt) \leq \thr $. If this is the case, we find, using the method of Lagrange multipliers, a distribution which satisfies the risk constraint and minimizes the squared distance from $ \tilde{\stepdist_\tstep} $; such a distribution is then output by \texttt{RiskAwareExplore}. 

If we needed to relax $ \thr $ to solve $ \linprog $, we assume the predictions to be too pessimistic and opt for a more radical exploration. Hence, we ignore $ \linprog $ altogether and instead select actions proportionally to their UCT values, i.e. we put $ \stepdist_\tstep(a) = \texttt{UCT}(\trroot(\tree),a)/\sum_{b\in \act}\texttt{UCT}(\trroot(\tree),b).$

\subsection{Predictor \& Training}
\label{subsec:training}

In principle any predictor (e.g. a neural net) can be used with \ralph{}. In this paper, as a proof of concept, we use a simple table predictor, directly storing the estimates for each state $ s $ (the parameter $ \theta $ can then be identified with the table, i.e. $\param(s)=\pred_{\param}(s)$). 


Each episode produces a data element $\epi = (s_0,\stepdist_0,\steprew_0)\cdots(s_{\hor-1},\stepdist_{\hor-1},\steprew_{\hor-1})$, where $s_\tstep$, $\stepdist_\tstep$, $\steprew_\tstep$ are the current state, the distribution on actions used, and the reward obtained in step $\tstep$, respectively. For every step $\tstep$ of this episode we compute the discounted accumulated payoff $G_\epi^{ \tstep} = \sum_{j=\tstep}^{\hor-1} \discount^{\tstep-j}\cdot \steprew_j$ from that step on; similarly, for risk  we set $ R_\epi^\tstep$ to $1$ if some $s_j \in \fail_\mdp$ for $j \geq \tstep$, and to $ 0 $ otherwise; for action probabilities we denote $ P_\epi^\tstep = \stepdist_i $. For each state $s$ encountered on $ \epi $ we put $I_\epi(s) = \{\tstep \mid 0\leq \tstep \leq H-1 \wedge s_\tstep = s\}$.

The state statistics across all episodes in $ \trset $ are gathered in an every-visit fashion~\cite{SB:book}. I.e., we compute the quantities $N(s) = \sum_{\eta\in \trset} |I_\epi(s)|$ (the total visit count of $ s $), $ G(s) = \sum_{\eta\in \trset,\tstep \in I_\epi(s)} G_\epi^{ \tstep} $, $ R(s) =  \sum_{\eta\in \trset,\tstep \in I_\epi(s)} R_\epi^{ \tstep}$, and $P(s) = \sum_{\eta\in \trset,\tstep \in I_\epi(s)} P_\epi^{ \tstep}$. These are then averaged to $ \widetilde{G}(s) = G(s)/N(s) $, $\widetilde{R}(s) = R(s)/N(s)$, and $ \widetilde{P}(s) = P(s)/N(s) $ (operations on probability distributions are componentwise). Together, these averages form a target table $\widetilde{\theta}$ such that $\widetilde{\theta}(s) = (\widetilde{G}(s),\widetilde{R}(s),\widetilde{P}(s))$. Finally, we perform the update $\theta \leftarrow \theta + \lrate(\widetilde{\param}-\param)$, where $\lrate$ is a pre-set learning rate.

This scheme can be generalized to more sophisticated predictors, which only requires replacing the final update with a gradient descent in the parameter space. The implementation and evaluation of these predictors is left for future work.

\section{Experiments}

\begin{figure}[]
\begin{center}
 \begin{BVerbatim}
 1 1 1 1 1 1
 1 A B x   1
 1 D C E g 1
 1 1 1 1 1 1
 \end{BVerbatim}
\end{center}
\caption{Example of a Hallway MDP.  Symbols '1', 'x', 'g' represent wall/trap/gold cell respectively; the other symbols are empty cells. The agent starts in B facing east.}
\label{fig:spinning}
\end{figure}

\begingroup
\begin{table*}[!t]
\makegapedcells
\begin{tabular}{|c|c|c|c|c|c|c|c|c|c|c|r|}
\hline
 & 
Algo & 
$\thr$ & 
$\expnodes$ & 
\multicolumn{1}{c|}{\begin{tabular}[c]{@{}c@{}}Avg\\ payoff\end{tabular}} & \multicolumn{1}{c|}{\begin{tabular}[c]{@{}c@{}}Stdev\\ payoff\end{tabular}} & \multicolumn{1}{c|}{\begin{tabular}[c]{@{}c@{}}Risk\end{tabular}} & \multicolumn{1}{c|}{\begin{tabular}[c]{@{}c@{}}Succ \\ avg \\ payoff\end{tabular}} & \multicolumn{1}{c|}{\begin{tabular}[c]{@{}c@{}}Succ\\ stdev\\ payoff\end{tabular}} & 
\multicolumn{1}{c|}{\begin{tabular}[c]{@{}c@{}}Training \\ time{[}s{]}\end{tabular}} &
\multicolumn{1}{c|}{\begin{tabular}[c]{@{}c@{}}Time per\\ episode\\ (avg){[}ms{]}\end{tabular}} & \multicolumn{1}{c|}{\begin{tabular}[c]{@{}c@{}}Total node \\ expansions\end{tabular}} \\ \hline\hline
\multirow{2}{*}{H 1} & \multirow{2}{*}{RAMCP} & 0 & 25 & 12.79 & 23.93 & 0.0 & 12.79 & 29.93 & N/A & 252.3  & 23,837,630 \\ \cline{3-12}
 &  & 0.1 & 25 & 30.78 & 34.77 & 0.082 & 35.76 & 31.74 &  N/A & 174.7  & 16,219,205 \\ \cline{3-12}
 &  & 0.25 & 25 & 45.08 & 36.86 & 0.193 & 61.80 & 14.43 &  N/A & 80.8 & 8,405,505  \\ \cline{2-12}
 & \multirow{2}{*}{\ralph{}} & 0 & 25 & 40 & 0.0 & 0.0 & 40 & 0 & 1.8 & 7.8  & 120,830  \\ \cline{3-12}
 &  & 0.1 & 25 & 46.78 & 25.80 & 0.094 & 53.70 & 14.95 &  1.2 & 5.9  & 96,748 \\ \cline{3-12}
 &  & 0.25 & 25 & 52.36 & 35.74 & 0.196 & 70 & 0.0 &  1.6 & 3.3  & 27,054 \\ \hline\hline
\multirow{2}{*}{H 2} & \multirow{2}{*}{RAMCP} & 0 & 50 & N/A & N/A & N/A & N/A  & N/A & \multicolumn{1}{c|}{N/A} & \multicolumn{1}{c|}{Timeout}  & \multicolumn{1}{c|}{N/A} \\ \cline{3-12}
 &  & 0.1 & 50 & N/A & N/A & N/A & N/A & N/A & N/A & \multicolumn{1}{c|}{Timeout}  & \multicolumn{1}{c|}{N/A} \\ \cline{3-12}
 
 &  & 1 & 50 & 60.15 & 39.27 & 0.15 & 73.57 & 24.44 &  N/A & 134,534  & 123,943,098 \\ \cline{2-12}
 & \multirow{2}{*}{\ralph{}} & 0 & 50 & 61.0 & 0 & 0.0 & 61.0 & 0.0 & 103 & 137  & 29,739,512 \\ \cline{3-12}
 & & 0.1 & 50 & 65.75 & 26.11 & 0.075 & 72.28 & 12.79 &  65  & 90 &  18,317,217 \\ \cline{3-12}
 & & 1 & 50 & 70.11 & 32.25 & 0.136 & 82.84 & 3.11 &  5 & 6  &  8,625,983  \\ \hline \hline
\multirow{2}{*}{H 3}  & \multirow{2}{*}{RAMCP} & 0 & 100 & 92.53 & 137.02 & 0.651 & 278.67 & 6.83 &  N/A & 294  & 138,800,592 \\ \cline{3-12}
 &  & 0.1 & 100 & 93.18 & 137.24 & 0.649 & 279.19 & 6.90 &  N/A & 287  &  145,421,231 \\ \cline{3-12}
 &  & 1 & 100 & 21.71 & 83.49 & 0.906 & 285.70 & 2.24 &  N/A & 59 & 59,666,989 \\ \cline{3-12}
 &  & 0.1 & {\bf 500} & 161.84 & 142.50 & 0.411 & 280.62 & 5.92 &  N/A & 1,582 &  651,055,909  \\ \cline{2-12}
 & \multirow{2}{*}{\ralph{}} & 0 & 100 & 281.169 & 5.02 & 0.0 & 281.169 & 5.02  & 16  & 108  &  8,069,542  \\ \cline{3-12}
 & & 0.1 & 100 & 281.723 & 9.51 & 0.001 & 281.169 & 5.02 &  73  & 154  &  45,766,309 \\ \cline{3-12}
 & & 1 & 100 & 280.00 & 20.80 & 0.005 & 281.46 & 2.72 &  16  & 26 &  42,912,980 \\ \cline{3-12}
 & & 0.1 & {\bf 50} & 279.32 & 29.14 & 0.01 & 282.24 & 2.41 &  8 & 63 &  3,733,503  \\ \cline{3-12}
 \hline\hline
\multirow{2}{*}{H 4}& \multirow{2}{*}{\ralph{}} & 0 & 50 & 1270.0 & 0.0 & 0.0 & 1270.0 & 0.0 & 631 & 903 & 107,845,003 \\ \cline{3-12}
 & & 0.1 & 50 & 1311.11 & 149.57 & 0.062 & 1349.54 & 2.09 &  821 & 1,034 & 123,007,131 \\ \cline{3-12}
 & & 1 & 50 & 1276.45 & 130.61 & 0.110 & 1307.06 & 11.10 &  26 & 36 & 53,565,317 \\ \hline \hline
 \multirow{2}{*}{RW 1}& RAMCP & 0.05 & 50 & 17.78 & 19.38 & 0.234 & 27.03 & 10.75 & N/A & 548 & 77,713,072 \\ \cline{2-12}
 & \ralph{} & 0.05 & 50 & 23.32 & 13.55 & 0.035 & 24.92 & 10.69 & 34 & 68 & 23,077,790 \\ \hline \hline 
 \multirow{2}{*}{RW 2}& RAMCP &  0.05 & 50 & \multicolumn{1}{c|}{N/A} & \multicolumn{1}{c|}{N/A} & \multicolumn{1}{c|}{N/A} & \multicolumn{1}{c|}{N/A} & \multicolumn{1}{c|}{N/A} & N/A & {Timeout} & \multicolumn{1}{c|}{N/A} \\ \cline{2-12}
 & \ralph{} & 0.05 & 50 & 97.33 & 24.38 & 0.007 & 98.18 & 22.24 &{62} & 113 & \multicolumn{1}{l|}{104,871,114} \\ \hline
\end{tabular}
\caption{Summary of the Hallway benchmark. Here H 1,...,4 correspond to Hallway 1,...,4, respectively.  RW 1 is a random walk with 50 states, RW 2 with 200 states. Parameter $\expnodes$ denotes the number of simulations per step.}
\label{table:benchmark_01}
\end{table*}
\endgroup

\mypar{Benchmarks} We implemented \ralph{} and evaluated it on two sets of benchmarks. The first one is a modified, perfectly observable version of Hallway \cite{PGT03,SS04} where we control a robot navigating a grid maze using three possible moves: forward, turn right, turn left. Depending on the instance, the forward movement might be subject to random perturbations (the robot shifted to the right or left of the target cell). For every step the robot incurs a (fixed) negative penalty. Some cells of the maze contain ``gold,'' collection of which yields a positive reward. Cells may also contain traps. Entering a trap entails a small chance of destroying the robot (i.e. going to a failure state). Each gold piece can be only collected once, so each additional gold cell doubles the size of the state space.

As a second benchmark, we consider a controllable \emph{random walk (RW)}. The state space here are integers in a fixed $0$-containing interval, representing the agent's wealth. At each step, the agent can chose between two actions - safer and riskier. Each action has a probabilistic outcome: either the wealth is increased or lost. The riskier action has higher expected wealth gain, but greater chance of loss. We start with a small positive wealth, and the failure states are those where the wealth is lost, i.e. non-positive numbers. In each step, the agent receives a reward/penalty equal to wealth gained/lost. The goal is to surpass a given wealth level $L$ as fast as possible: the agent incurs a small penalty for every step up to the first step when she surpasses~$ L $.



\mypar{Comparison} For comparison, we reimplemented the online RAMCP algorithm from \cite{CENR:ijcai18} slightly modified (per suggestion in the source paper) so as to allow for state-based risk. This should allow us to evaluate the effect of RAlph's crucial features (learning and prediction, risk-averse exploration) on the performance. To get a fair comparison, our implementation of RAMCP shares as much common code with \ralph{} as possible. 
In particular, both algorithms employ UCT-like simulations. We denote by $\simnum$ the number of these simulations invoked per decision. 

\mypar{Evaluation}
We evaluate \ralph{} and RAMCP on four instances of the Hallway (called Hallway 1, 2, 3, 4) of dimensions 2x3, 3x5, 5x8, 5x5. The corresponding MDPs have state-spaces of sizes $\lvert \states \rvert$ equal to 20, 44, 1136, 6553600, respectively. 
For the random walk, we consider benchmarks with $50$ and $200$ wealth levels (i.e. states).

The test configuration was: CPU: Intel Xeon E5-2620 v2@2.1GHz (24 cores); 8GB heap size; Debian 8. A training phase of \ralph{} is executed on 23 parallel threads, evaluation is single-threaded. Both algorithms were evaluated over 1000 episodes, with a timeout of 1 hour per evaluation. \footnote{Implementation can be found at \url{https://github.com/snurkabill/MasterThesis/releases/tag/AAAI_release}}

\mypar{Metrics} For both RAMCP and \ralph{}, we report the average payoff and risk. To account for bias caused by runs that ended in failure, we also consider payoff averaged over runs that avoided a failure state (``Succ avg payoff'' in Table~\ref{table:benchmark_01}). We also measured the training time of \ralph{} and, for both algorithms, an average time per evaluation episode. We also use the \textit{total node expansion} metric, tracking the number of search tree nodes created throughout the whole experiment on a given benchmark. For \ralph{}, this includes both training and evaluation; hence it is a relevant indicator of how much searching both methods require to produce the  results.

\begin{figure}[!t]
    \centering
    \includegraphics[width=\columnwidth]{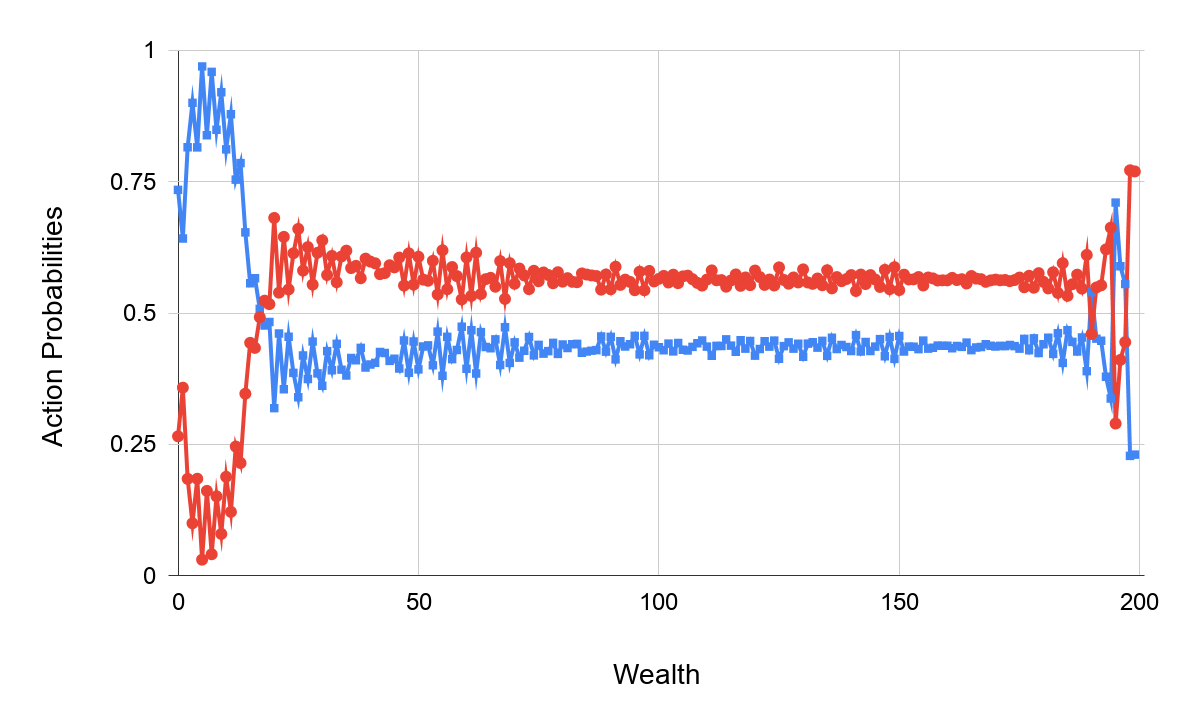}
    \caption{Action probabilities learned by \ralph{} for each wealth level of the RW benchmark. Blue line with boxes - safe action; red line with circles - unsafe action.}
    \label{fig:random_walk_policy}
\end{figure}

\mypar{Results}
The results are summarized in Table~\ref{table:benchmark_01}.
Even in smaller benchmarks, \ralph{} is much faster and makes up to two orders of magnitude less node expansions. 
This is because RAMCP lacks the knowledge that \ralph{} acquires during the training phase and thus RAMCP often keeps hitting walls or blunder in circles. Also, \ralph's risk-averse exploration improves the chance of finding promising paths. Although the learning is an advantage to \ralph{}, the total number of node expansions (including the learning phase) is much smaller than in RAMCP, which tends to construct large search trees. In Hallway 3, the average payoff of solutions found by RAMCP are inferior to those found by \ralph{} in approximately half of the time; and while the failure-avoiding runs of RAMCP perform similarly to those of \ralph{}, RAMCP is not able to consistently avoid failures and its risk is well above $\thr$ (the same holds for the RW benchmark). The reason is that RAMCP is too slow to find a competitive solution in the given time limit.
Enlarging the number of expanded nodes in every step  ($\expnodes$) of RAMCP is not sufficient to beat \ralph{}. On the other hand, changing $\expnodes$ from $100$ to $50$ in \ralph{} does not have a significant effect on solution quality. The results for Hallway 4 show that \ralph{} scales well for larger state spaces. RAMCP is omitted for Hallway 4, as each of its executions timed out.


%


\mypar{Discussion}
We observed an interesting connection between \ralph{} and AlphaZero. 
The behavior of \ralph{} with $\thr = 1$ is close in nature to the behavior of AlphaZero.
 If \ralph{} is invoked with $\thr$ small or zero, it explores the state space much faster (measured by node expansion count) than with $\thr=1$. The reason is that the risk-averse exploration of \ralph{} typically visits much smaller part of the state-space. Hence, in cases when risky paths are sub-optimal, \ralph{} may find a solution faster than algorithms ignoring the risk.

\ralph{} also exhibited interesting behavior on the Hallway instance shown in Figure~\ref{fig:spinning}. For $\thr = 0$, the only way to reach the gold is by exploiting the move perturbations: since the robot cannot to move east from C without risking a shift to the trap, it must keep circling through A, B, C, D until it is randomly shifted to E. \ralph{} is able, with some parameter tuning, to find this policy.

In the random walk benchmark, \ralph{} finds a common-sense solution of playing the safe action when the wealth is low and the riskier one otherwise. Figure~\ref{fig:random_walk_policy} depicts the probabilities of choosing the respective actions at all wealth levels (up to the level $L=200$). The differences of the probabilities for larger levels ($\geq 50$) are due to the step penalty equal to $-1$. For a larger penalty, the difference would be larger as the agent would be motivated to reach the top level $L$ faster. The wiggliness for wealths close to $L$ is caused by the specific structure of the optimal strategy. Indeed, for some specific wealth values close to $L$ it is beneficial to take the safer action, and \ralph{} exploits this peculiarity.


\section{Conclusions \& Future Work}
We introduced \ralph{}, an online algorithm for risk-constrained MDPs. Our experiments show that even with a simple predictor, \ralph{} performs and scales significantly better than a state-of-the-art algorithm. As an interesting future work we see extension of the method to POMDPs and incorporation of more sophisticated predictors.

\section*{Acknowledgements}
Krishnendu Chatterjee is supported by the Austrian Science Fund (FWF) NFN Grant No. S11407-N23 (RiSE/SHiNE), and COST Action GAMENET. Tom\'a\v{s} Br\'azdil is supported by the Grant Agency of Masaryk University grant no. MUNI/G/0739/2017 and by the Czech Science Foundation grant No. 18-11193S. Petr Novotn\'y and Ji\v{r}\'i Vahala are supported by the Czech Science Foundation grant No. GJ19-15134Y.

\bibliography{aaai20,krish-intro,new}
\bibliographystyle{aaai}
\end{document}